\documentclass[accepted]{tpm2023}
\usepackage[british]{babel}
\usepackage{natbib} 
\usepackage{mathtools} 
\usepackage{booktabs} 
\usepackage{tikz,pgfplots,bm,algorithm,algorithmic} 
\pgfplotsset{compat=1.17} 
\usetikzlibrary{pgfplots.statistics,arrows,plotmarks,decorations.markings,trees,shapes,pgfplots.groupplots,automata,circuits}
\tikzset{dot/.style = {circle, fill, minimum size=#1,inner sep=0pt, outer sep=0pt, fill, circle},dot/.default = 6pt}
\tikzset{dot2/.style = {circle, fill, color=black!40,minimum size=6pt,inner sep=0pt, outer sep=0pt, fill, circle}}
\tikzstyle{a}=[->,>=stealth,dashed]
\tikzstyle{a2}=[->,>=stealth]
\tikzstyle{nodo}=[ellipse,draw=black!100,fill=black!0,line width=.7pt,minimum width=1.2cm,minimum height=0.8cm,text width=1.2cm,text centered]
\tikzstyle{nodo2}=[ellipse,draw=black!100,fill=black!10,line width=.7pt,minimum width=1.2cm,minimum height=0.8cm,text width=1.2cm,text centered]
\tikzstyle{nodo3}=[ellipse,draw=black!100,fill=black!30,line width=.7pt,minimum width=1.2cm,minimum height=0.8cm,text width=1.2cm,text centered]
\tikzstyle{arco}=[draw=black!80,line width=.7pt, postaction={decorate}, decoration={markings,mark=at position 1.0 with {\arrow[ draw=black!80,line width=.7pt]{>}}}]
\title{Tractable Bounding of Counterfactual Queries by Knowledge Compilation}
\author[1]{\href{mailto:<david.huber@idsia.ch>?Subject=Your TPM 2023 paper}{David Huber}{}}
\author[2]{Yizuo Chen}
\author[1]{Alessandro Antonucci}
\author[2]{Adnan Darwiche}
\author[1]{Marco Zaffalon}
\affil[1]{%
    IDSIA\\
    Lugano, Switzerland}
\affil[2]{%
    UCLA\\
    Los Angeles, US}
\begin{document}
\maketitle
\begin{abstract}
We discuss the problem of bounding partially identifiable queries, such as counterfactuals, in Pearlian structural causal models. A recently proposed iterated EM scheme yields an inner approximation of those bounds by sampling the initialisation parameters. Such a method requires multiple (Bayesian network) queries over models sharing the same structural equations and topology, but different exogenous probabilities. This setup makes a compilation of the underlying model to an \emph{arithmetic circuit} advantageous, thus inducing a sizeable inferential speed-up. We show how a single \emph{symbolic} knowledge compilation allows us to obtain the circuit structure with symbolic parameters to be replaced by their actual values when computing the different queries. We also discuss parallelisation techniques to further speed up the bound computation. Experiments against standard Bayesian network inference show clear computational  advantages with up to an order of magnitude of speed-up.
\end{abstract}

\section{Introduction}\label{sec:intro}
Causal inference is an important direction for modern AI. Following Pearl's \emph{ladder of causation} \citep{ladder}, observational data are sufficient to compute correlational queries, while answering interventional queries requires an additional structure such as the causal graph and dedicated computational schemes such as the popular \emph{do calculus} \citep{pearl2009causality}. Moving further into counterfactual inference requires the full specification of the underlying causal model, including the structural equations and the exogenous parameters. While the equations might be available (or sampled), the exogenous parameters are typically latent and unavailable. Most counterfactuals are therefore \emph{partially identifiable} and only bounds are obtained for the corresponding queries \citep{shpitser2012counterfactuals}.

Despite the hardness of the task \citep{zaffalon2021}, approximate bounding  schemes exist. These include polynomial programming \citep{duarte2021}, credal networks inference \citep{zaffalon2020}, sampling \citep{zhang2021counterfactual}, and EM \citep{zaffalon2021}. The latter, in particular, allows us to derive credible intervals while reducing the bounds' computation to iterated (Bayesian network) inferences in a fully specified structural causal model. Such a method requires multiple queries over models sharing the same structural equations but different exogenous probabilities. 

Tractable \emph{arithmetic circuits} (e.g., \citet{darwiche2022tractable}) offer a graphical formalism to represent generative probabilistic models and compute standard inferential tasks in linear time by a circuit traversal. The ACE library\footnote{\url{http://reasoning.cs.ucla.edu/ace}.} allows Bayesian network compilation to arithmetic circuits with state-of-the-art performances \citep{agrawal2021partition}.

The goal of this paper is to adopt the above compilation strategy to achieve a sizeable inferential speed-up in the computation of bounds for counterfactual queries. In particular, we consider a \emph{symbolic} knowledge compilation as in \citet{darwiche2022causal} to obtain the circuit structure with the symbolic parameters to be replaced by their actual values when computing the different queries (Sect.~\ref{sec:compilation}). We also present parallelisation techniques to further speed up the bound computation. Experiments based on ACE against standard Bayesian network algorithms report computational speed-ups up to an order of magnitude (Sect.~\ref{sec:experiments}). This contribution appears to be the first application of knowledge compilation to counterfactual inference. A discussion on the outlooks of these strategies is in Sect.~\ref{sec:conclusions}.


\section{Notation and Basics}\label{sec:background}
Variable $X$ takes values from a finite set $\Omega_X$, $\theta_X$ is a probability mass function (PMF) over $X$, $\theta_x$ denote the probability of $X=x$, and $\lambda_x$ the indicator function of that event.

\begin{figure}[htp!]
\centering
\begin{tikzpicture}[scale=1]
\node[] ()  at (-1,0.5) {\scriptsize $\theta_{X_1}=[0.1,0.9]$};
\node[] ()  at (3,1) {\scriptsize $\theta_{X_2|X_1=0}=[0.2,0.8]$};
\node[] ()  at (3,0.5) {\scriptsize $\theta_{X_2|X_1=1}=[0.3,0.7]$};
\node[dot,label=below left:{$X_1$}] (v1)  at (0,0) {};
\node[dot,label=below right:{$X_2$}] (v2)  at (1.6,0) {};
\draw[a2] (v1) -- (v2);
\end{tikzpicture}
\caption{A BN over two Boolean variables.}\label{fig:bn}
\end{figure}
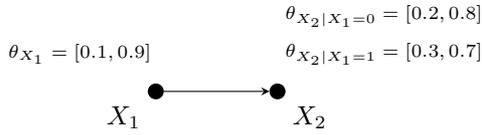

\paragraph{Bayesian Networks (BNs)} 
Given variables $Y$ and $X$, a conditional probability table (CPT) $\theta_{Y|X}$ is a collection of PMFs over $Y$ indexed by the values of $X$. Given a joint variable $\bm{X}:=(X_1,\ldots,X_n)$ and a directed acyclic graph $\mathcal{G}$ with nodes  in a one-to-one correspondence with the variables in $\bm{X}$, a BN is a collection of CPTs ${\bm{\theta}}:=\{\theta_{X_i|\mathrm{Pa}_{X_i}}\}_{i=1}^n$, where $\mathrm{Pa}_{X_i}$ denotes the \emph{parents} of $X_i$ according to $\mathcal{G}$ (see, e.g., Fig.~\ref{fig:bn}). A BN induces a PMF $\theta_{\bm{X}}$ s.t. $\theta_{\bm{x}}=\prod_{i=1}^n \theta_{x_i|\mathrm{pa}_{X_i}}$, for each $\bm{x}\in\Omega_{\bm{X}}$. 

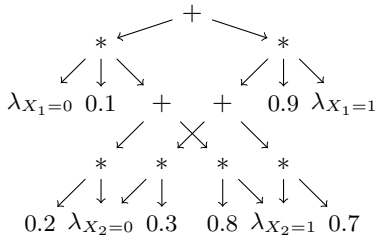
\begin{figure}[htp!]
\centering
\begin{tikzpicture}[scale=0.8]
\node[] (l11)  at (0,1) {$+$};
\node[] (l21)  at (-1.5,0.5) {$*$};
\node[] (l22)  at (1.5,0.5) {$*$};
\node[] (l31)  at (-2.5,-0.5) {\footnotesize $\lambda_{X_1=0}$};
\node[] (l32)  at (-1.5,-0.5) {\footnotesize $0.1$};
\node[] (l33)  at (-0.5,-0.5) {$+$};
\node[] (l31b)  at (0.5,-0.5) {$+$};
\node[] (l32b)  at (1.5,-0.5) {\footnotesize $0.9$};
\node[] (l33b)  at (2.5,-0.5) {\footnotesize $\lambda_{X_1=1}$};
\node[] (p1)  at (-1.5,-1.5) {$*$};
\node[] (p2)  at (-0.5,-1.5) {$*$};
\node[] (p3)  at (0.5,-1.5) {$*$};
\node[] (p4)  at (1.5,-1.5) {$*$};
\node[] (l1)  at (-2.5,-2.5) {\footnotesize $0.2$};
\node[] (l2)  at (-1.5,-2.5) {\footnotesize $\lambda_{X_2=0}$};
\node[] (l3)  at (-0.5,-2.5) {\footnotesize $0.3$};
\node[] (l4)  at (0.5,-2.5) {\footnotesize $0.8$};
\node[] (l5)  at (1.5,-2.5) {\footnotesize $\lambda_{X_2=1}$};
\node[] (l6)  at (2.5,-2.5) {\footnotesize $0.7$};
\draw[->] (l11) -- (l21);
\draw[->] (l11) -- (l22);
\draw[->] (l21) -- (l31);
\draw[->] (l21) -- (l32);
\draw[->] (l21) -- (l33);
\draw[->] (l22) -- (l31b);
\draw[->] (l22) -- (l32b);
\draw[->] (l22) -- (l33b);
\draw[->] (l33) -- (p1);
\draw[->] (l33) -- (p3);
\draw[->] (l31b) -- (p2);
\draw[->] (l31b) -- (p4);
\draw[->] (p1) -- (l1);
\draw[->] (p1) -- (l2);
\draw[->] (p2) -- (l2);
\draw[->] (p2) -- (l3);
\draw[->] (p3) -- (l4);
\draw[->] (p3) -- (l5);
\draw[->] (p4) -- (l5);
\draw[->] (p4) -- (l6);
\end{tikzpicture}
\caption{An AC implementing the BN in Fig.~\ref{fig:bn}.}\label{fig:ac}
\end{figure}

\paragraph{Arithmetic Circuits (ACs)}
We can express the joint PMF of a BN as a multi-linear function of the CPT parameters, i.e., 
$\theta_{\bm{x}} = \sum_{x_1',\ldots,x_n'} \prod_i \theta_{x_i'|\mathrm{pa}_{X_i}} \lambda_{x_i}$. Such an exponential-size representation becomes more compact by exploiting the BN conditional independence relations induced by $\mathcal{G}$ and consequently moving the sums inside the products. The representation might be even more compact if different CPT parameters take the same value. Common examples are context-specific independence relations and CPTs implement deterministic relations through \emph{degenerate} (i.e., $0$/$1$) values only. Such functions are graphically depicted as ACs composed by leaves, annotated by CPT probabilities and indicator functions, and inner nodes containing sums and multiplications (e.g., Fig.~\ref{fig:ac}). Those ACs are called \emph{tractable}, as they allow to answer some queries in linear-time, through feed-forward passes on the circuit structure. A number of \emph{compilation} algorithms have been proposed to build compact AC representations of BNs.

\begin{figure}[htp!]
\centering
\begin{tikzpicture}[scale=0.8]
\node[] ()  at (-2.1,1.5) {\scriptsize $\theta_{U_1}=[0.1,0.9]$};
\node[] ()  at (-2.1,0.2) {\scriptsize $f_{V_1}(U_1)=U_1$};
\node[] ()  at (4.5,1.5) {\scriptsize $\theta_{U_2}=[0.05,0.15,0.25,0.55]$};
\node[] ()  at (4.5,0.2) {\scriptsize $f_{V_2}(V_1,U_2=\left[\begin{array}{c}0\\1\\2\\3\end{array}\right])=\left[\begin{array}{c}0\\V_1\\\neg V_1\\1\end{array}\right]$};
\node[dot2,label=above left:{$U_1$}] (u1)  at (0,1) {};
\node[dot2,label=above right:{$U_2$}] (u2)  at (1.6,1) {};
\node[dot,label=below left:{$V_1$}] (v1)  at (0,0) {};
\node[dot,label=below right:{$V_2$}] (v2)  at (1.6,0) {};
\draw[a] (u1) -- (v1);
\draw[a] (u2) -- (v2);
\draw[a2] (v1) -- (v2);
\end{tikzpicture}
\caption{A FSCM over two endogenous (black) and two exogenous variables (grey nodes).}\label{fig:scm}
\end{figure}
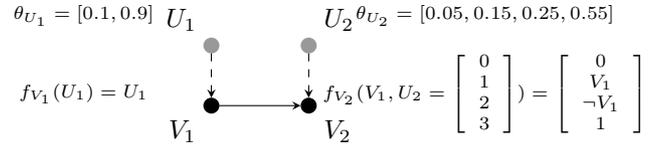

\paragraph{Structural Causal Models (SCMs)} A \emph{structural equation} (SE) $f$ associated with variable $Y$ and based on the input variable(s) $X$, is a surjective function $f:\Omega_{X} \to \Omega_Y$ that determines the value of $Y$ from that of $X$.  Given two joint variables $\bm{U}$ and $\bm{V}$, called respectively \emph{exogenous} and \emph{endogenous}, a collection of SEs $\{f_V\}_{V\in\bm{V}}$ such that, for each $V\in\bm{V}$ the input variables of $f_V$ are in $(\bm{U},\bm{V})$, is called a \emph{partially specified} SCM (PSCM). A PSCM induces a directed graph $\mathcal{G}$ with nodes in correspondence with the variables in $(\bm{U},\bm{V})$ and such that there is an arc between two variables if and only if the first variable is an input variable for the SE of the second (e.g., Fig.~\ref{fig:scm}). We focus on \emph{semi-Markovian} PSCMs, i.e., those PSCMs that lead to acyclic graphs. A \emph{fully specified} SCM (FSCM) is just a PSCM $M$ paired with a collection of marginal PMFs, one for each exogenous variable. As SEs induce (degenerate) CPTs, an FSCM defines a BN over $(\bm{U},\bm{V})$ based on $\mathcal{G}$.

\paragraph{Causal Queries in FSCMs} BN algorithms allow to compute inferences in FSCMs. This is trivially the case for observational queries involving joint or conditional states of the endogenous variables. For interventional queries, this can be also done provided that the SEs of the intervened variables are replaced by constant maps pointing to the selected state. For counterfactual queries, where the same variable may be observed as well as subject to intervention, albeit in distinct \emph{worlds}, we use auxiliary structures where different copies of the endogenous variables and their SEs are considered in each world. \citet{han2023complexity} provides a precise characterisation of the computational complexity of those inferences in terms of treewidth.

\paragraph{Partially Identifiable Causal Queries in PSCMs} FSCMs are rarely available. Considering a PSCM specification with a dataset $\mathcal{D}$ of endogenous observations represents a more common setup. This is not critical for observational queries: a BN over the endogenous variables can be obtained by deriving its graph from that of the PSCM and the CPTs from $\mathcal{D}$ \citep{tian2002studies}. Interventional queries can be possibly reduced to observational queries by the \emph{do calculus} \citep{pearl2009causality}. If this is not possible, we say that the query is only \emph{partially identifiable}. In those cases, a characterisation is still provided by the bounds spanned by the values of the query computed for all the FSCMs consistent with the PSCM and the endogenous BN (e.g., \citet{zaffalon2020}). Counterfactual queries are very often only partially identifiable.


\section{Tractable Bounding of Counterfactuals}\label{sec:compilation}
Bounding partially identifiable queries PSCMs is an NP-hard problem even on polytrees \cite[Theorem~2]{zaffalon2021}. 

\citet{zhang2021counterfactual} have proposed a Bayesian sampling procedure that eventually approximates the bounds via credible intervals. The sampling is query-driven; new queries will require new sampling. The accuracy of the approximation is unclear in general as a systematic experimental analysis is missing.

\paragraph{EM Approach} The algorithm proposed by \cite{zaffalon2021} samples the initialisation of the exogenous chances, which are used to start an EM scheme returning a compatible FSCM specification. Alg.~\ref{alg:cem} depicts a single EM run. The interval spanned by the values of the query computed on the FSCMs returned by the EM for each run provides an (inner) approximation of the expectation bounds. This approach is `agnostic' w.r.t. the query. It aims at reconstructing the uncertainty related to exogenous variables (via sets of probabilities). Once this is done, different (counterfactual) queries will use the same sets of probabilities to compute the wanted bounds---no more sampling is needed.

\begin{algorithm}[htp!]
\caption{In a PSCM paired with an endogenous dataset $\mathcal{D}$, given a random initialisation $\{\theta_U^{(0)}\}_{U\in\bm{U}}$ in input, the algorithm returns the exogenous chances $\{\theta_U\}_{U\in\bm{U}}$ obtained after likelihood convergence.}
\begin{algorithmic}[1]
\STATE $t \leftarrow 0$
\WHILE{$P(\mathcal{D}|\{\theta_{U}^{t+1}\}_{U\in\bm{U}}) \geq P(\mathcal{D}|\{\theta_{U}^{t}\}_{U\in\bm{U}})$}
\FOR{$U\in \bm{U}$}
\STATE $\theta^{t+1}_U \leftarrow |\mathcal{D}|^{-1}\sum_{\bm{v} \in \mathcal{D}} \theta^t_{U|\bm{v}}$
\STATE $t \leftarrow t+1$
\ENDFOR
\ENDWHILE
\end{algorithmic}\label{alg:cem}
\end{algorithm}

An approximate bounding scheme based on  Alg.~\ref{alg:cem} may suffer from two potential bottlenecks: (i) an insufficient number of runs leading to a poor inner bound approximation; (ii) the time needed by the FSCM inferences required by the exogenous queries (line 4) and the likelihood evaluation (line 2). 

Regarding (i), \cite{zaffalon2022} derived a characterisation of the accuracy of the bounds in terms of credible intervals, and the EM scheme has been proven to yield accurate bounds with relatively few runs. 

Here we address instead (ii) by first noticing that the queries needed by Alg.~\ref{alg:cem} are computed on different FSCMs based on the same PSCM, thus having possibly different exogenous chances, but always the same endogenous CPTs implementing the SEs of the PSCM. This is true for the models corresponding to different time steps $t$, but also when different exogenous initialisations are considered in input. In practice the algorithm requires the computation of inferences in different BNs having the same CPTs for the non-root nodes, but different marginal PMFs on the root nodes. This simple remark suggests the use of AC compilation to achieve faster inferences.

\paragraph{Symbolic Knowledge Compilation} 
Consider the AC compilation of two BNs over the same variables and with the same graph but different CPT parameters. Suppose these parameters, separately for each BN, have no repeated values. In that case, the compiler minimises the size of the ACs by only exploiting the independence relations induced by the BN graph. As these are the same for the two BNs, the two ACs returned by the compiler should share the same inner nodes and the same indicators on the leaves while differing only on the chances in the leaves. 

This fact allows for a \emph{symbolic} compilation achieved by regarding the chances in the leaves as symbolic parameters to be replaced by their actual values during an inferential computation. Compilers can quickly implement symbolic compilation by replacing the BN parameters with unique numerical identifiers to be eventually retrieved in the AC returned by the compiler.

Returning to the queries of interest for the EM scheme, we might intend PSCM compilation as a symbolic compilation achieved by treating the exogenous PMFs as parameters. In contrast, the endogenous CPTs, implementing the SEs and remaining the same for all the models, are treated as constant numerical values. The degenerate nature of the CPTs can be exploited by the compiler to achieve smaller ACs and hence faster inferences (e.g., with the FSCM in Fig.~\ref{fig:scm} as input, ACE returns an AC with 96 arcs if the determinism of the CPTs is not exploited and 23 arcs otherwise). After the PSCM symbolic compilation, the AC of each FSCM required by Alg.~\ref{alg:cem} is obtained in linear time (w.r.t. the AC size) by replacing the parameters of the \emph{symbolic} AC with the actual values in the particular FSCM. 

The queries required by Alg.~\ref{alg:cem} are, for each $\bm{v}\in\mathcal{D}$, the computation of endogenous marginal $\theta_{\bm{v}}$ and the exogenous posterior $\theta_{u|\bm{v}}$, to be computed for each $U\in\bm{U}$ and $u\in\Omega_U$. We therefore focus on the computation of the joint query $\theta_{u,\bm{v}}$ for each $U\in\bm{U}$, $u\in\Omega_U$ and $\bm{v}\in\mathcal{D}$. This is performed in linear time by a bottom-up traversal of the AC after instantiating the indicators of the variables in $\bm{V}$ and $U$.

\paragraph{Parallelisation} 
Alg.~\ref{alg:cem} allows for a straightforward parallelisation at the run level. A more sophisticated parallelisation can be based on \emph{c-components} \citep{tian2002studies}. In a PSCM, a c-component is a set of variables connected through undirected paths consisting solely of exogenous-to-endogenous arcs. For each c-component, we define a sub-graph consisting of the nodes in the c-component and its direct parents, with all other variables and edges removed. The corresponding sub-model might yield the chances of the exogenous variables in the c-component through Alg.~\ref{alg:cem}. The procedure can be executed in parallel, separately for each c-component. 


\section{Experiments}\label{sec:experiments} 
To evaluate the benefits of the proposed AC approach when running the EM scheme in Alg.~\ref{alg:cem}, we compare the AC execution times against those based on standard BN inference for a synthetic benchmark of $335$ PSCMs. The PSCM graphs have a random topology (Erd\"os-R\'enyi sampling), the number of nodes ranges between $5$ and $21$ (avg. $9.9$), and the number of root nodes (i.e., exogenous variables) between $2$ and $10$ (avg.~$4.5$). All the endogenous variables are binary, while the cardinalities of the exogenous ones range between $3$ and $256$ (avg.~$29.6$). Each PSCM comes with a dataset of endogenous observations of size between $1,000$ and $5,000$ records obtained by sampling a compatible FSCM. The benchmark and the code used for the simulations are available in a dedicated repository.\footnote{\url{anonymous.4open.science/r/uai-E5D7}.}

The code is built on the top of CREDICI\footnote{\url{github.com/idsia/credici}.} \citep{credici}, a Java library implementing the EM scheme and embedding a BN inference engine. Here we consider inferences based on variable elimination with the min-fill heuristics. The symbolic compilation is instead developed within the Java/C++ ACE compiler (see Footnote~1). The experiments are run on a dual 2.20GHz Intel(R) Xeon(R) Silver 4214 CPU Dell PowerEdge R540 server running Ubuntu 20.04.6 LTS. All the experiments are performed using a fixed seed for the random initialisation and, as expected, resulted in the exact same set of PSCMs. 

For each PSCM, we perform $200$ runs of $500$ iterations. We set a timeout to $15$ minutes for each experiment. The BN approach based on the whole model often reaches this limit. Thus, as a baseline for the BN approach, we consider the faster BNC approach based on queries in the sub-BNs associated with the model c-components. The number of c-components for the benchmark models ranges between $1$ and $10$ (avg.~$4.2$). The parallelisation of BNC over the different components is denoted instead as BNP. We similarly denote as ACC the method based on the (symbolic) compilation of the sub-BNs and as ACP its parallelisation. The overall execution times (in hours) on the whole benchmark for the four methods are $T_\mathrm{BNC}=17.0$, $T_\mathrm{BNP}=7.3$, $T_\mathrm{ACC}=2.4$, and $T_\mathrm{ACP}=1.3$. This clearly shows the advantage of the (symbolic) knowledge compilation.

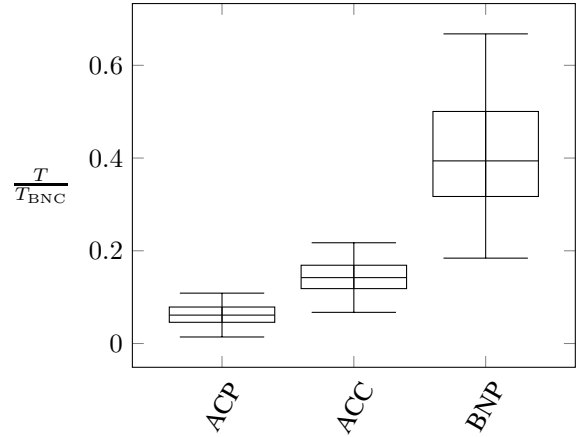
\begin{figure}[htp!]
\centering
\begin{tikzpicture}[scale=1]
\begin{axis}
    [
    boxplot/draw direction = y,
    cycle list={{black}},
    x post scale=0.85,
    y post scale=0.85,
    xticklabel style = {align=center, rotate=60},
    ylabel style = {align=center, rotate=-90},
    ylabel = {$\frac{T}{T_{\mathrm{BNC}}}$},
    xtick={1,2,3},
    xticklabels={ACP, ACC, BNP}
]
\addplot+[
    boxplot prepared={
      median=0.061309726024659776,
      upper quartile=0.07872098344329165,
      lower quartile=0.045784983496369516,
      upper whisker=0.014133122921087043,
      lower whisker=0.10866572006368057
    },
    ] coordinates {};
    
\addplot+[
    boxplot prepared={
      median=0.14214068001249588,
      upper quartile=0.1688512362243524,
      lower quartile=0.11844673196938087,
      upper whisker=0.06725456016224038,
      lower whisker=0.2173202941824813
    },
    ] coordinates {};
    
\addplot+[
    boxplot prepared={
      median=0.3938830665378548,
      upper quartile=0.5004657274637524,
      lower quartile=0.31711804761846263,
      upper whisker=0.1841618148162964,
      lower whisker=0.6677853792866055
    },
    ] coordinates {};
  \end{axis}
\end{tikzpicture}
\caption{Runtime savings w.r.t. BNC.}
\label{fig:exp}
\end{figure}

A deeper analysis is provided by computing, separately for each PSCM, the ratio between the EM execution time of a particular approach and that of BNC. Fig.~\ref{fig:exp} shows the boxplots of the different approaches. In practice, using ACs makes the bounding of the counterfactual queries one order of magnitude faster. 
Note also that we considered PSCM of bounded size ($\leq 21$ nodes) just to permit a comparison against the BN approaches, which cannot handle bigger networks in reasonable time limits.

\section{Conclusions}\label{sec:conclusions}
In this study, we have investigated the potential of knowledge compilation within the framework of partially identifiable queries, such as counterfactuals, in structural causal models. We have assumed that structural equations are given together with a dataset of endogenous observations. From these we reconstruct the uncertainty about the exogenous variables with sets of probabilities.

The advantages of using knowledge compilation appear clear: the new approach leads to one order of magnitude speed-up compared to pre-existing models based on Bayesian nets.

As future work we intend to use the knowledge compilation approach to execute the EM scheme in very large models along two dimensions: the size of the network as well as the cardinality of exogenous variables. The latter is in particular an important factor to represent general `canonical' specifications of PSCMs. These specifications enable one to be dispensed of the requisite to provide structural equations in input: a causal graph with endogenous data would suffice to compute counterfactual inference.

We also intend to explore more in-depth problems with network structures with large treewidth that may thus be intractable by variable elimination.

\bibliography{references}
\end{document}